\documentclass[letterpaper,journal]{IEEEtran}
\IEEEoverridecommandlockouts
\usepackage[utf8]{inputenc}
\usepackage{amssymb}
\usepackage{arydshln}
\usepackage{booktabs}
\usepackage[bookmarks=true,hidelinks]{hyperref}
\usepackage{cite}
\usepackage{amsmath,amssymb,amsfonts}
\usepackage{algorithmic}
\usepackage{cleveref}
\usepackage{graphicx}
\usepackage{textcomp}
\usepackage{orcidlink} 
\usepackage{booktabs}
\usepackage{upgreek}
\usepackage[section]{placeins}
\usepackage{url}
\usepackage{tabularx}
\newcommand{\bettershortstack}[2][c]{\begin{tabular}[c]{@{}#1@{}}#2\end{tabular}}

\title{\LARGE \bf
Towards Autonomous Soft Robotic Endovascular Navigation via Imitation Learning
}
\author{Noah Barnes$^{1}$, Ji Woong Kim$^{2}$, Lingyun Di$^{3}$, Hannah Qu$^{1}$, Anuruddha Bhattacharjee$^{1}$, Miroslaw Janowski$^{4}$, \\Dheeraj Gandhi$^{4}$, Bailey Felix$^{6}$, Shaopeng Jiang$^{5}$, Olivia Young$^{6}$, Mark Fuge$^{7}$, Ryan D. Sochol$^{6}$, \\Jeremy D. Brown$^{1}$, and Axel Krieger$^{1}$
\thanks{$^{1}$Johns Hopkins University, \{nbarne18, hqu6, abhatt27, jdelainebrown, axel\}@jhu.edu}%
\thanks{$^{2}$Stanford University, jwbkim@stanford.edu} 
\thanks{$^{3}$McGill University, lingyun.di@mail.mcgill.ca}
\thanks{$^{4}$University of Maryland, Baltimore, \{miroslaw.janowski, dheeraj.gandhi\}@som.umaryland.edu}
\thanks{$^{5}$Swiss Federal Institute of Technology in Lausanne (EPFL), shaopeng.jiang@epfl.ch}
\thanks{$^{6}$University of Maryland, College Park, \{bmfelix, oyoung, rsochol\}@umd.edu}
\thanks{$^{7}$ETH Zurich, mafuge@ethz.ch}%
\thanks{This work was supported in part by National Institutes of Health R01EB033354. In addition, the work was supported in part by the Maryland Robotics Center and the Center for Engineering Concepts Development at the University of Maryland. Finally, this material is based upon work supported by the National Science Foundation Graduate Research Fellowship Program under Grant No. DGE 2236417 and 2139757. Any opinions, findings, and conclusions or recommendations expressed in this material are those of the author(s) and do not necessarily reflect the views of the National Science Foundation.}
}

\begin{document}
\maketitle
\thispagestyle{empty}
\pagestyle{empty}
\begin{minipage}{\textwidth}
\vspace{-2.5em}
\centering
\textbf{This work has been submitted to the IEEE for possible publication. Copyright may be transferred without notice, after which this version may no longer be accessible.}
\vspace{2em}
\end{minipage}
\begin{abstract}
In endovascular surgery, endovascular interventionists push a thin tube called a catheter, guided by a thin wire to a treatment site inside the patient's blood vessels to treat various conditions such as blood clots, aneurysms, and malformations. Robotic guidewires can enhance maneuverability but are difficult to model and control. Autonomous soft robotic guidewire navigation has the potential to overcome these challenges, increasing the precision and safety of endovascular navigation. As a first step, we establish a large-scale, 2D-projected environment for autonomous navigation. In other surgical domains, end-to-end imitation learning has shown promising results. Thus, we develop a transformer-based imitation learning framework with goal conditioning, relative action outputs, and automatic contrast dye injections to enable generalizable soft robot navigation in an aneurysm targeting task. We train the policy on 36 different modular bifurcated geometries, generating 647 total demonstrations under simulated fluoroscopy, and evaluate it on three previously unseen vascular geometries. The policy reaches the aneurysm with a success rate of 83\% on the unseen geometries, outperforming several baselines. In addition, ablation and baseline studies evaluate the effectiveness of each design and data collection choice. Lastly, we extend the policy to achieve 75\% success on an unseen patient-derived geometry. Project website: https://softrobotnavigation.github.io/
\end{abstract}


\section{Introduction}
\begin{figure}[ht]
\begin{center}
\centerline{\includegraphics[width=0.8\linewidth]{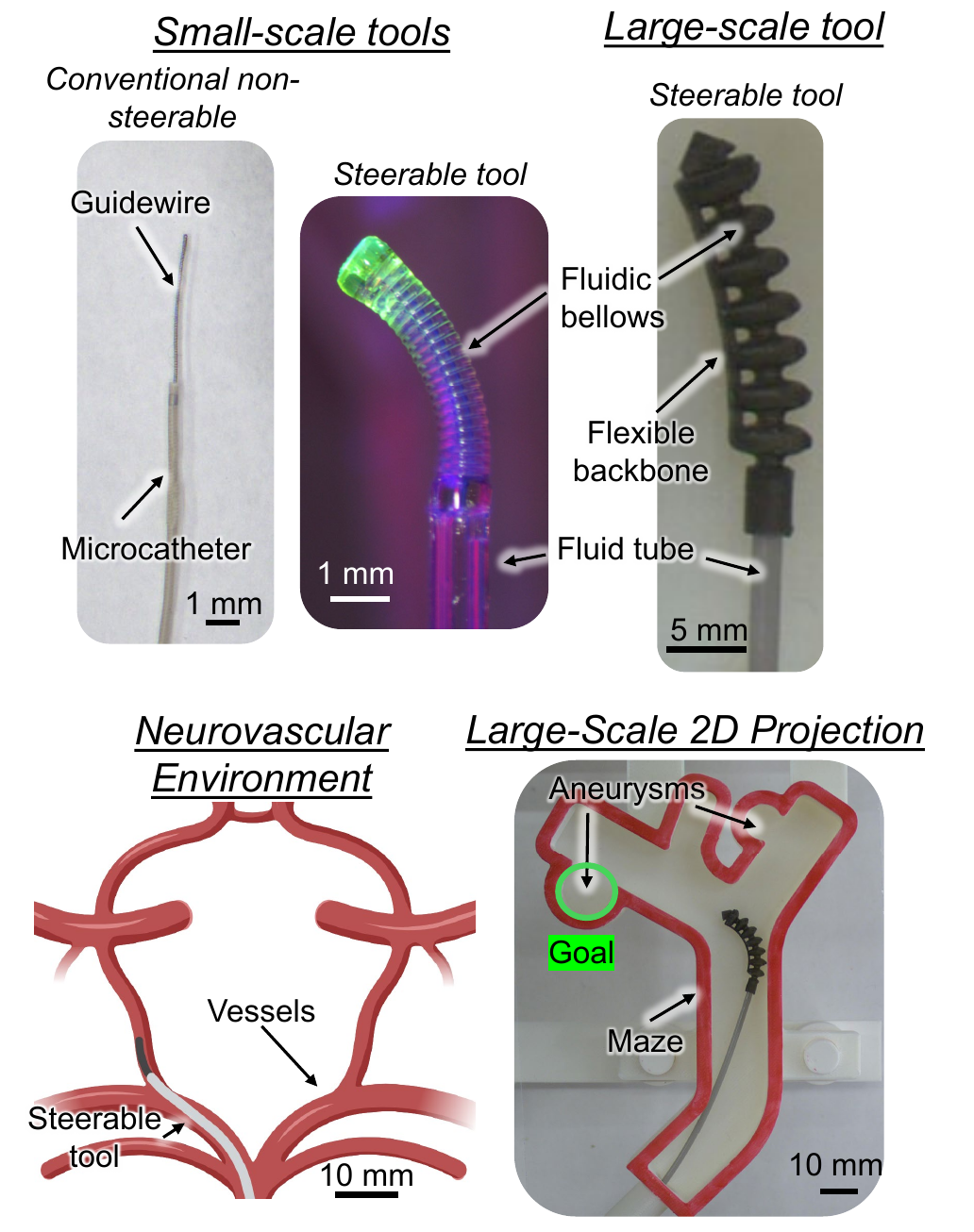}}
\caption{\textit{(Top)} Commercial guidewire and microcatheter for neurovascular intervention next to a soft robotic microcatheter~\cite{felixsoromicrocath} and our tool. \textit{(Bottom)} Illustration of a soft robotic tool inside the vessels in the Circle of Willis (neurovascular structure). Here, we deploy a 3D-printed soft robotic tool in a 2D projection of various vascular geometries.} 
\label{fig:abs}
\end{center}
\vspace{-1\baselineskip}
\end{figure}

\IEEEPARstart{D}{iagnosis} and treatment of vascular conditions require an endovascular interventionist to skillfully advance catheters and guidewires through the patient's blood vessels. Robotically steerable tools can improve maneuverability over conventional tools~\cite{Cruddas2021}. However, complex vessel-tool forces stemming from the severe under-actuation and infinite degrees-of-freedom of the flexible tools still prohibit a consistent mapping between the physician's actions outside of the body and the tool's movement inside the body~\cite{Pore2023}. Autonomy can improve precision, reduce complications and procedural times, and limit fluoroscopic radiation exposure on the patient and operating team~\cite{Santiago2024}. To this end, we investigate an autonomous navigation approach for soft robotic guidewires and catheters. 

In addition to the persistent control challenges, there are significant visual constraints in endovascular surgery. Under X-ray fluoroscopy, the vessels are only visible after a radiopaque contrast injection,  which dissipates within seconds. Once the vessels fill, a snapshot of the vessels is captured. This static vessel roadmap can be referenced while navigating the guidewire, but it is only an approximation due to constant vessel deformation and patient movement. Moreover, sensorization remains challenging due to the tools' millimeter-to-sub-millimeter sizes and high flexibility~\cite{Ren2023}. Even if adequate tool localization within the blood vessels is achieved, navigation and control are still non-trivial. Interventionists rely on trial-and-error involving a combination of advancing, retracting, and rotating to enter the device into the correct vessel~\cite{Santiago2024}.

In this work, the application of interest is intracranial aneurysm treatment, in which a neurointerventional surgeon navigates a microguidewire and microcatheter (\textless 1 mm diameters) through the blood vessels of the brain to the site of an aneurysm to deliver therapy~\cite{Santiago2024}. In these cases, complex bifurcations, unusual aneurysm orientation, and tortuosity lead to procedural difficulty and complications. As a critical step towards achieving autonomous navigation, we conduct our study using a large-scale robot. While progress towards miniaturizing soft robotic tools is being made (see Fig. 1 and~\cite{felixsoromicrocath}), larger-scale prototypes are currently more mechanically robust and easier to track in camera images. Thus, they enable reliable collection of hundreds of demonstrations before small-scale extension. In addition, we isolate our focus to a single bifurcation projected to a 2D plane. Clinically, navigation can be reduced to one roughly planar bifurcation at a time after C-arm positioning. The setup is depicted in Fig.~\ref{fig:abs}. With this setup, we preserve several core difficulties of endovascular intervention: unpredictable vessel-tool forces via a steerable soft robot attached to a flexible tube; high geometrical variation via modular 3D-printed vessel mazes; and ambiguous and incomplete visual feedback via a fluoroscopy simulator.

To address these difficulties, we introduce an end-to-end imitation learning framework that uses a transformer-based action-chunking policy. To enable generalization to varying geometries and goal locations, we construct a feature map that encodes the distance to the goal at each pixel in the vessel roadmap. This feature map undergoes a small rigid transformation to simulate the roadmap's inaccuracy. Further, we output the robot's motor position commands relative to the motor position at the time the policy is queried, since absolute motor positions do not have a consistent corresponding robot position. Finally, the policy is given control over when to inject contrast dye. The robot cannot always be localized within the vessels without a contrast injection, but contrast dye must be limited in practice due to its toxicity. As such, the policy effectively learns when a localization update is needed.

This work contributes a policy that: (i) is the first end-to-end imitation learning algorithm for soft robotic endovascular navigation, (ii) uses goal conditioning, relative actions, and learned contrast injections to improve performance under simulated fluoroscopy, and (iii) is evaluated through ablative and baseline experiments that verify the importance of each design choice for generalizing to unseen geometries on physical platforms. Sections~\ref{sec:genmethods}-\ref{sec:genresult} describe our generalized bifurcated geometry navigation, and Sections~\ref{sec:neuromethods}-\ref{sec:neuroresults} extend the framework to a patient-derived neurointerventional case study.

\section{Related Work}
\label{sec:relatedwork}
Soft robot control approaches include online Jacobian estimation~\cite{Yip2014}, physics and finite-element based modeling~\cite{Roshanfar2024, Thieffry2019}, and learning-based controllers~\cite{Fang2019, Wu2022}. However, these works do not address the effects of friction, energy buildup, distributed contact, and wall-induced tool deformation~\cite{Chen2024}. Thus, model-based~\cite{Ravigopal2021} catheter and guidewire navigation algorithms have been explored. However, they can be computationally expensive in complex anatomy and often omit soft robot dynamics. Naive strategies such as wall-following~\cite{Fagogenis2019} and center-line following~\cite{Li2024Centerline} show promise but require accurate reconstruction of the anatomy or on-robot vision~\cite{Pore2023}.

Learning-based approaches may be well-suited to emulate interventionists' trial-and-error behavior. A popular approach is reinforcement learning (RL), which trains an agent by exploration to maximize a cumulative reward, often defined in navigation tasks as reaching a target location. In many RL-based studies~\cite{Robertshaw2026MT, Moosa2025, Karstensen2024,Scarponi2024}, policy performance is either not fully evaluated on physical platforms or is significantly worse on physical platforms than in simulation. Imitation learning (IL) instead trains an agent to replicate the behavior of an expert demonstrator, rather than relying on exploration. Some research that uses RL or a combination of IL and RL~\cite{Yao2025, Chi2020} presents notable successful demonstrations on physical phantoms. However, Chi et al.~\cite{Chi2020} rely on electromagnetic-based 3D catheter tip tracking and offer limited generalization to new anatomies, while Yao et al.~\cite{Yao2025} execute an open-loop policy trained on a digital-twin simulation, limiting robustness to anatomical mismatch or unexpected intraoperative changes.

There have been limited attempts to achieve autonomous tool navigation through pure IL. For example, Zhao et al.~\cite{Zhao2022} trained a generative adversarial network (GAN) for aortic arch path planning, and Peloso et al.~\cite{Peloso2025} used behavioral cloning for mitral valve path planning; neither addresses dexterous tool manipulation. End-to-end IL has succeeded in other surgical domains~\cite{Kim2025}, but has not been adapted to endovascular challenges, including multiple potential goal locations that necessitate goal-conditioning and the coupled sensing-action problem of contrast management. 

Further, explicit goal-conditioning, which is unaddressed by previous end-to-end IL surgical policies, is required to generalize to arbitrarily-shaped vessels. Outside of surgery, goal-conditioned IL strategies specify a task objective in a variety of ways, including images of the goal state in robotic manipulation~\cite{Sundaresan2025}, and waypoints in autonomous driving~\cite{Chen2019}. Our goal map is related to distance-transform inputs used in vision models~\cite{MaDistSeg2020}, as well as potential-field, value-map~\cite{TamarValue2016}, and waypoint-heatmap~\cite{Krantz2021ICCV} navigation. However, we do not use the map as an explicit trajectory or model-based controller. Instead, the goal map conditions an end-to-end policy, allowing the robot to learn local maneuvers such as leveraging wall contact, retraction, and recovery from overshoot.

\section{Methods for Generalized Navigation}
\label{sec:genmethods}
\subsection{Hardware Setup}
We used a 3D-printed fluid-driven soft robot with a 5 mm diameter bellows actuator similar to~\cite{Jiang2024}, driven by a syringe pump for bending and a compact (7$\times$8$\times$8.5 cm) belt-driven translation stage. Fluid-driven actuation is an attractive choice for steerable catheters and guidewires~\cite{felixsoromicrocath} due to their inherent compliance and simple control infrastructure. Demonstrations were collected using the teleoperated force handle from~\cite{Barnes2024}, which maps user-applied forces to bending and translation velocities. The control software combines LabVIEW front-end with Python-based image processing and policy inference. The platform provides a reliable physical testbed for investigating autonomous navigation, but the proposed learning framework is not dependent on this particular hardware design.
\begin{figure}[ht]
\begin{center}
\centerline{\includegraphics[width=\linewidth]{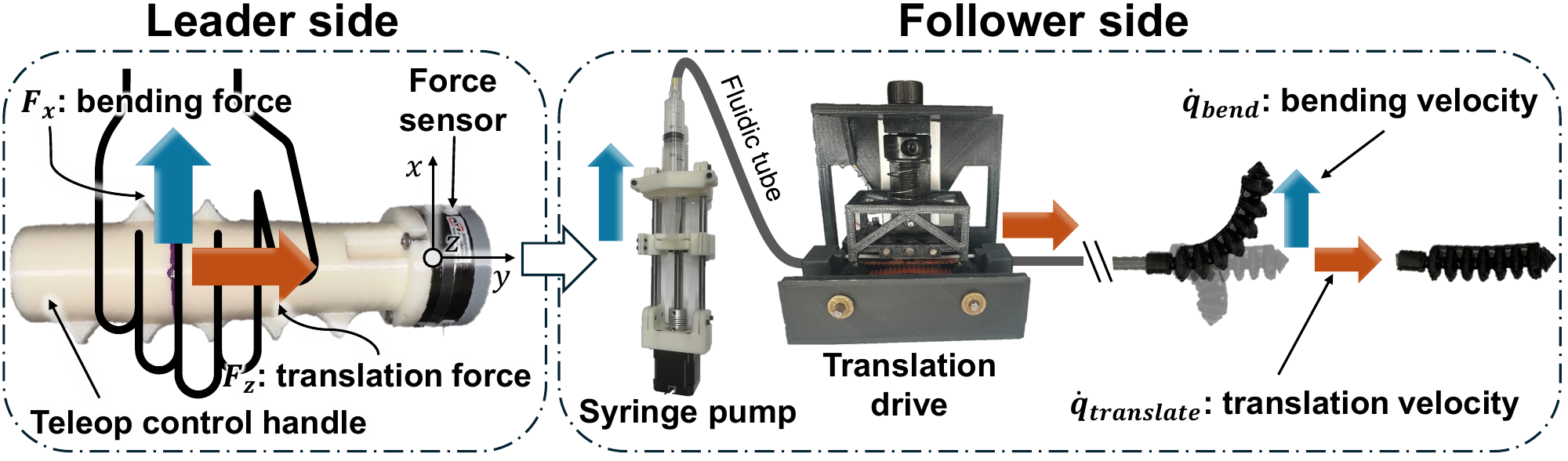}}
\caption{To control the soft robot, a user inputs force commands through a teleoperated control handle. These forces are proportionally mapped to the bending and translation velocities of the robot, achieved by the syringe pump and translation drive, respectively.} 
\label{fig:hardwarefig}
\end{center}
\vspace{-1\baselineskip}
\end{figure}

\subsection{Experimental setup}
\begin{figure}[t]
\begin{center}
\centerline{\includegraphics[width=\linewidth]{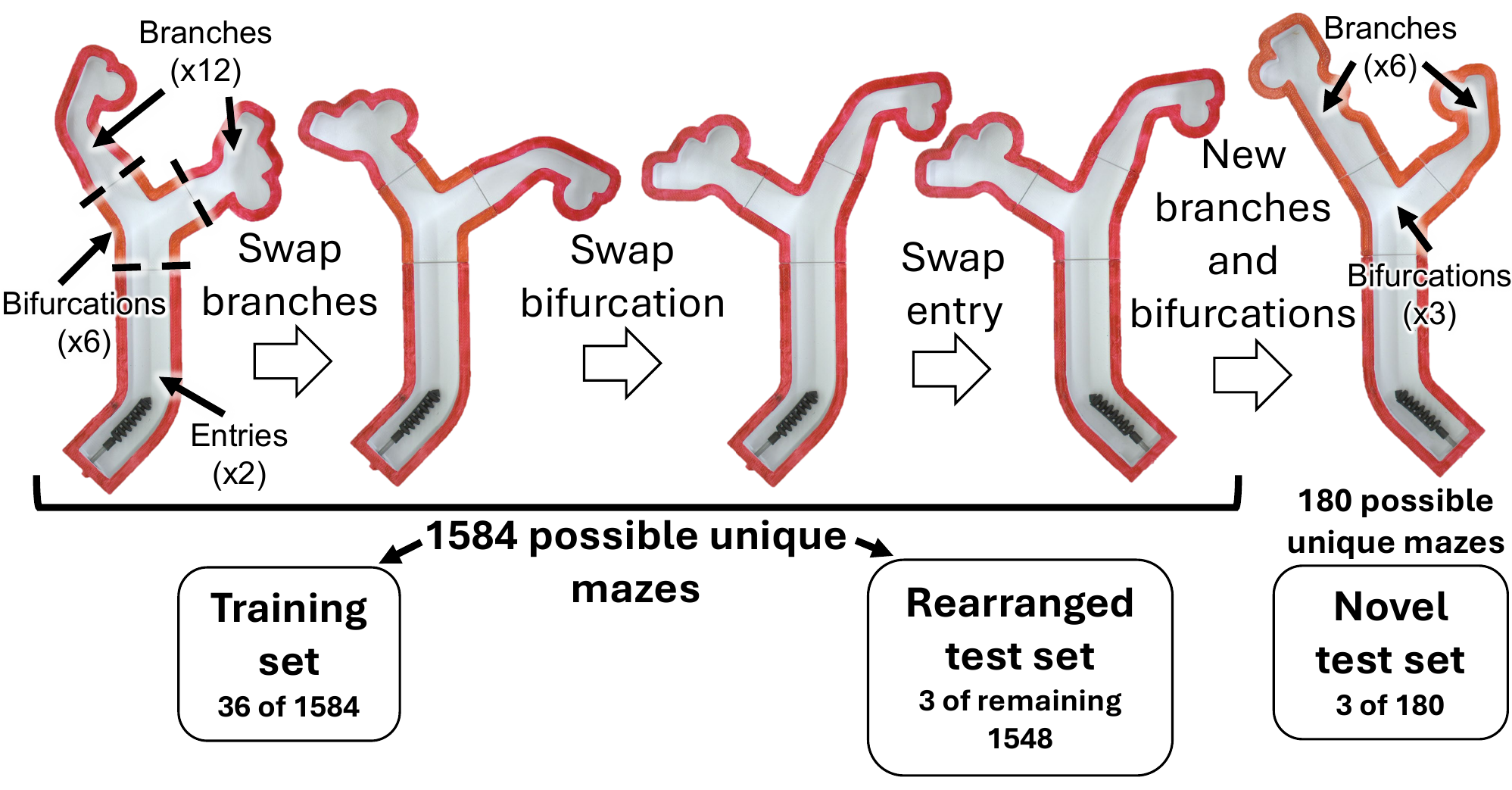}}
\caption{First, we reserve a certain set of bifurcations and branches for the training set. From these sets of blocks, we choose a subset of combinations for the training set and a different subset for the rearranged test set. The novel test set is formed by a new set of bifurcations and branches.} 
\label{fig:mazesetup}
\end{center}
\vspace{-1\baselineskip}
\end{figure}

\textbf{Maze design.} To simulate an aneurysm navigation task in a 2D environment, we designed modular 3D-printed mazes with interchangeable entries, bifurcations, and branches. The bifurcations varied in the angle of each connecting branch across the range of 25-70 degrees. The branches varied in width (8-16 mm), aneurysm distance from the bifurcation (45-65 mm), aneurysm diameter (5-13 mm), and which side the aneurysm branches from. Additional variations included secondary bends, empty branches, and bumps along the wall. The training set used 36 mazes assembled from one set of modules. The rearranged test set used three new combinations of the modules that appeared in training, whereas the novel test set used 3 mazes composed entirely of new bifurcation and branch modules.  Thus, the novel geometries contain branches and bifurcations that did not appear in the training set in any configuration. This design explicitly evaluates two levels of generalization: transfer to unseen arrangements of familiar components (``rearranged"), and transfer to local geometries absent from the training set (``novel") (see Fig.~\ref{fig:mazesetup}). 

\textbf{Robot segmentation and tip tracking:} Robot segmentation and tip location were estimated in real time using a UNet~\cite{Ronneberger2015}, trained on ground-truth masks from Segment Anything Model 2 (SAM 2)~\cite{Ravi2024} and tip annotations from the skeletonized mask. The UNet was trained for 72 hours on an NVIDIA RTX 4000 GPU, with a dataset of 8,352 images from 28 videos. Manual annotation of 50 randomly selected images from the test trials yielded a median, 95th percentile, and maximum tip localization error of 0.6, 1.9, and 8.0 mm (2.0, 6.7, 28.8 px), respectively. While electromagnetic tracking could provide an alternative means of tip localization, it suffers from calibration, interference, and physical tool integration limitations~\cite{RAMADANI2022102584}. Tracking in the 2D image plane is sufficient for our task.

\begin{figure}[ht]
\begin{center}
\centerline{\includegraphics[width=\linewidth]{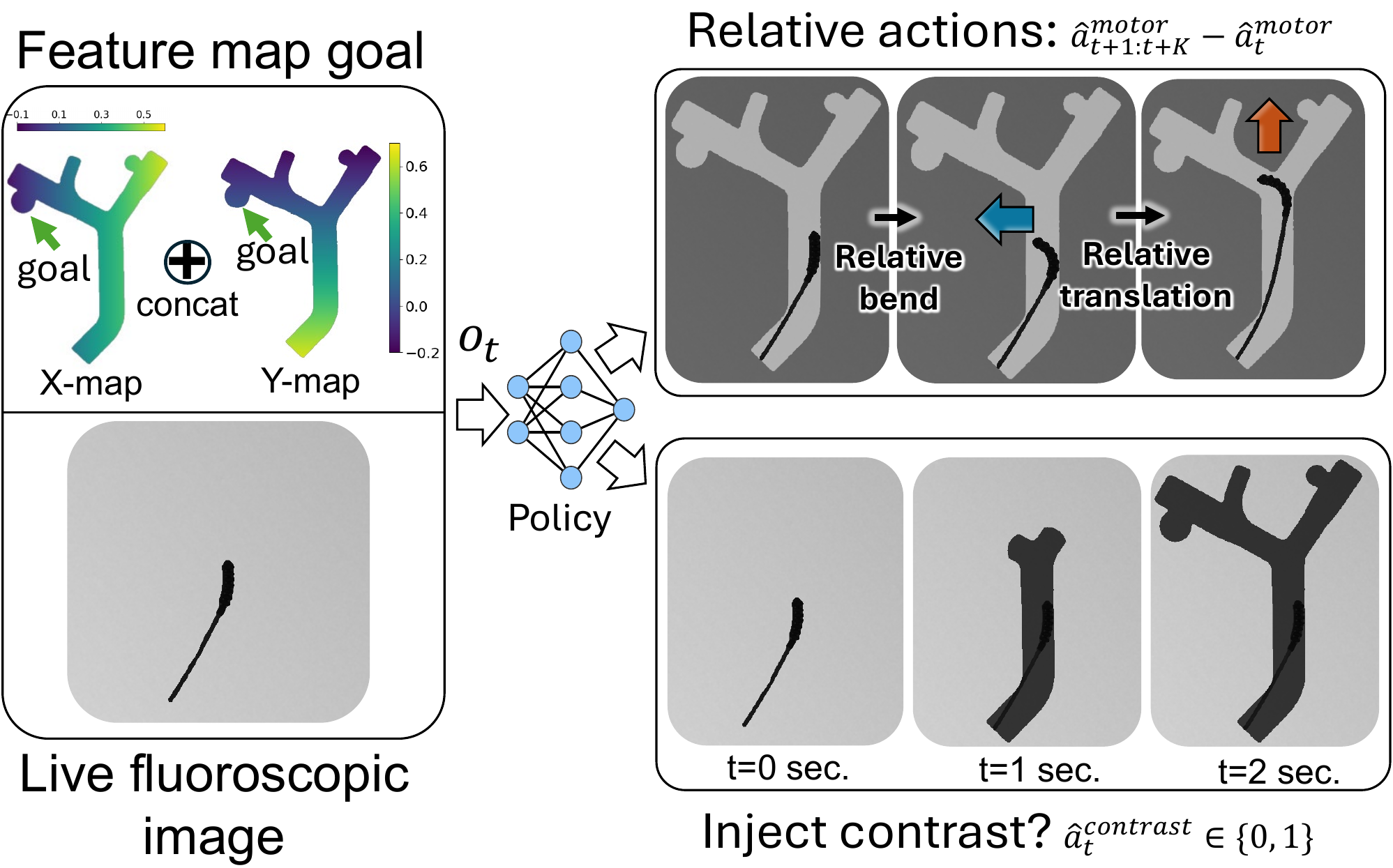}}
\caption{Proposed architecture for autonomous navigation. The static goal representation and live fluoroscopic image are passed to the autonomous policy. The policy outputs a sequence of relative actions and a binary variable indicating whether to inject contrast or not.} 
\label{fig:algofig}
\end{center}
\vspace{-1\baselineskip}
\end{figure}

\textbf{Fluoroscopy simulation:} To emulate realistic fluoroscopic feedback, we translated the overhead camera images, such as those shown in Fig.~\ref{fig:mazesetup}, into simulated fluoroscopic images, such as those in Fig.~\ref{fig:algofig}. The fluoroscopic view consists of the robot segmentation overlaid in black over a noisy light gray image. When a contrast injection is initiated, a dark gray color fills the vessels by following the path of a breadth-first search along the vessel segmentation (obtained via color thresholding) centerline.
The contrast fills the vessels over 1-2 seconds, depending on the size of the vessels, remains for four seconds, and then dissipates for 1-2 seconds.

\textbf{Data collection:} Initially, ten demonstrations were performed for each of the 36 mazes. In this initial set, we collected 218 normal demonstrations and 142 recovery demonstrations. During normal demonstrations, the robot starts at the base of the maze's entrance. In contrast, during recovery demonstrations, the robot begins in an expected failure mode (e.g., in the wrong branch, past the target, or nearly buckling against the wall). The experimenter could press a button to inject contrast at any point. Once the robot's tip point enters the area of the target aneurysm, the trial is considered successful, and the time elapsed is recorded. If, for a set of demonstrations on a particular target, the time elapsed is more than 60 seconds on average, the geometry was excluded. This criterion led to one set of branches being discarded due to the aneurysm being outside of the robot's achievable workspace. 

After training a policy on the 360 initial demonstrations, more data was generated according to the dataset aggregation (DAgger) method~\cite{Ross2011}: the policy was evaluated on all the training mazes, and when the policy failed, the evaluation was paused, and the experimenter teleoperated the robot to the target. Only the recovery segment was recorded. Thus, the policy can learn to recover from incorrect actions that it is prone to execute. By this method, 287 recovery demonstrations were added to the training set, resulting in 647 total demonstrations (218 normal, 429 recovery).
\vspace{-0.5\baselineskip}
\subsection{Policy Implementation}
We use a transformer-based action chunking policy, inspired by the Surgical Robot Transformer (SRT)~\cite{Kim2024}, which applies Action-Chunking Transformers~\cite{Zhao2023} to surgical tasks with the da Vinci robot. Action-chunking can combat compounding errors in long-horizon imitation learning tasks by predicting a sequence of future actions rather than a single action~\cite{Zhao2023}, a technique that is well suited to our task, in which using only intermittent visual updates has clinical benefits. The novelty is not action chunking itself, but its adaptation to endovascular navigation through vessel-based goal conditioning, relative motor commands, and joint contrast-injection prediction. 

The policy consists of a BERT~\cite{Devlin2019}-style transformer encoder-decoder pair with image feature-maps computed by Resnet-18 encoders~\cite{He2015} as its input token sequence. The policy learns the distribution:
\begin{align}
    \pi_{\theta}(a_{t+1:t+K}^{motor}-a_{t}^{motor},a_t^{contrast}|o_t)
\end{align}
Where $a_s^{motor}=(a_s^{bend}, a_s^{translate})$ are the motor positions at timestep $s$, $K$ is the chunk size, $a_t^{contrast}$ is a binary flag indicating whether a contrast injection was initiated at any time in the chunk, and $o_t$ is the image observation. 

The image observation consists of the live fluoroscopic image and the feature map goal representation. The goal representation is two grayscale images that are concatenated to form a $(512\times 612\times 2)$ image, and the live fluoroscopic image is a single grayscale $(512\times 612\times 1)$ image. Intuitively, the goal representation provides a guiding vector at each pixel location within the vessels. This provides richer information than an image of the goal state or location and is naturally suited for input to a CNN. We generate this representation by concatenating two images: one in which the pixel values represent the y-distance of their position to the center of the target aneurysm, and another for the x-distance. The pixels outside of the vessels are set to 1. The calculation of pixel value is as follows:
\begin{align}
    u_{i,j}&=\frac{i-x_a}{c}*m_{i,j}+(1-m_{i,j}),\\
    v_{i,j}&=\frac{j-y_a}{c}*m_{i,j}+(1-m_{i,j}),
\end{align}
where $u_{i, j}$ and $v_{i, j}$ represent the values in the $(i, j)$th index of feature maps $U$ and $V$, respectively, $m_{i, j}$ is the value of the $(i, j)$th index of the vessel mask in which the pixels inside the vessels have value 1 and outside have value 0, $x_a$ and $y_a$ are the x- and y-coordinates of the aneurysm center, and $c$ is a normalizing constant, chosen to scale the maximum possible distance given by the image's spatial dimensions to 1. In a real procedure, the vessels often move after the static vessel roadmap, from which the goal image is derived, is captured. Thus, we apply a small random rigid transformation (range of $\pm$3 mm, $\pm$3 deg.) to the goal image to simulate this inadvertent movement. The goal and fluoroscopic images are center-cropped and resized to $(224\times 224)$ and passed through their own respective Resnet encoders pretrained on ImageNet. Sinusoidal embeddings are added to the encoder outputs, and the sum is passed through a transformer encoder and decoder, which outputs the action chunk.

The motor actions are predicted relative to the motor positions at time $t$, which is useful for imprecise robot kinematics~\cite{Kim2024} and allows the policy to infer a local relationship between robot shape and syringe displacement despite hysteresis and nonlinear dynamics.

Since contrast injections are sparse and last several seconds, the policy predicts whether an injection is initiated within a chunk, avoiding severe time-step-level class imbalance. Thus, the contrast injection prediction for a given chunk can be formulated as:
\begin{align}\hat{a}^{contrast}_t=\sigma\left(\max_{s\in [t+1, t+K]}\hat{c}_s\right),
\end{align}
where $\hat{c}_s$ is a binary flag predicting whether contrast is injected at timestep $s$, and the sigmoid function $\sigma(\cdot)$ converts the output to a probability.

The policy is trained via a supervised imitation loss:
\begin{align*}
    L(\theta)=&\lambda_a ||\hat{a}_{t+1:t+K}^{motor} - (a_{t+1:t+K}^{motor}-a_{t}^{motor})||_1 \\
    &+ \lambda_cBCE_{\beta}(\hat{a}_t^{contrast}, a_t^{contrast})
\end{align*}
Where $\lambda_a$ and $\lambda_c$ are the weights for the L1 motor action loss and the weighted Binary Cross Entropy (BCE) loss, respectively. Positive class predictions in the BCE loss are assigned a weight of $\beta$. 

Let $D_j=\{(o_t, a_{t+1:t+K}^{motor}, a_t^{contrast})\}$ denote the demonstration dataset after round $j$. After training the policy on $D_j$, the policy is evaluated, and recovery segments are added to $D_j$ to form $D_{j+1}$. The policy is then retrained on the aggregated dataset, reducing covariate shift by adding demonstrations from states induced by the learned policy.

We used an AdamW optimizer with learning rate $5\times10^{-5}$ and weight decay $1\times10^{-4}$, batch size 64, and trained for 3000 epochs. We chose $\lambda_a=1.0$ and $\lambda_c=0.5$ such that the BCE loss reaches the same order of magnitude as the L1 loss after a few epochs. We chose $\beta=7$ to reflect the average frequency of contrast injections observed in the training data. During inference, contrast is injected if $\hat{a}^{contrast}_t>0.5$. The policy contains 95 million parameters, and inference can run at 25 frames per second (fps) on an NVIDIA RTX 4000 GPU. 
\section{Generalized Navigation Experiments and Results}
\label{sec:genresult}
\begin{table}[t]
    \caption{Success rates across the ablative policies for the rearranged and novel geometries, along with pairwise logistic regression significance levels vs. Ours (*p\textless0.05 **p\textless0.01 ***p\textless0.001)}
    \centering
    \begin{tabular}{c | c c}
    Policy & Rearranged geometries & Novel geometries\\
    \bottomrule
    \toprule
    \textbf{Ours} & \textbf{89\% (16/18)} & \textbf{83\% (15/18)}\\
    50\% recovery & \textbf{89\% (16/18)} & 61\% (11/18)\\
    0\% recovery & 39\%*** (7/18) & 22\%*** (4/18)\\
    16 sec contrast & 78\% (14/18) & 56\% (10/18)\\
    8 sec contrast & 50\%** (9/18) & 50\%** (9/18)\\
    Binary goal & 61\% (11/18) & 61\% (11/18)\\
    No goal & 39\%*** (7/18) & 33\%*** (6/18)\\
    Absolute actions & 72\% (13/18) & 39\%* (7/18)
    \end{tabular}
    \label{tab:successablate}
\end{table}
\begin{figure}[ht]
\begin{center}
\centerline{\includegraphics[width=0.95\linewidth]{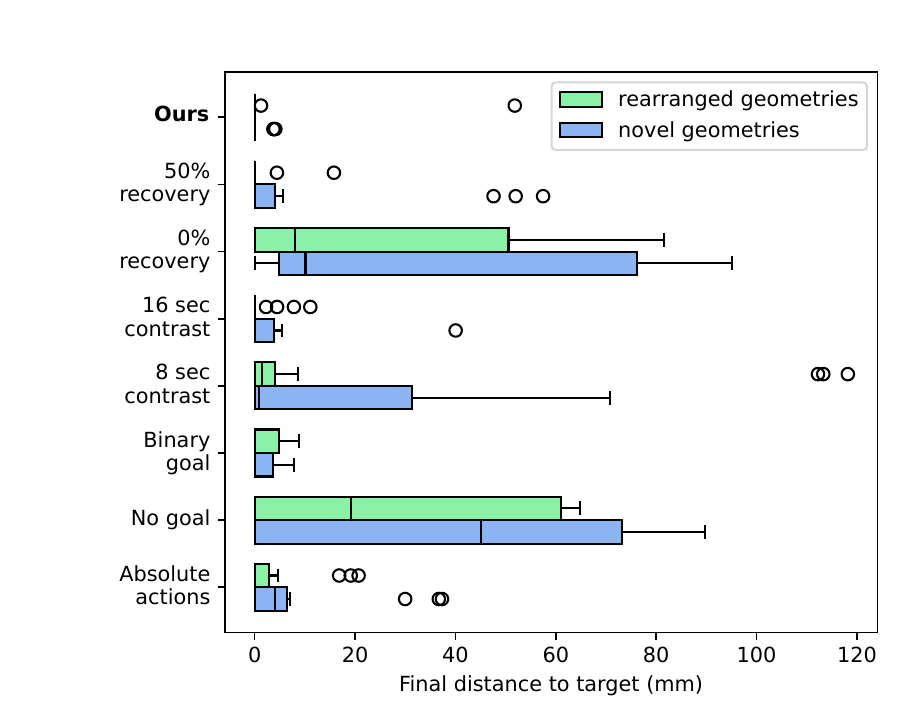}}
\caption{Distance to the goal aneurysm boundary at the end of the trial for each ablative policy. Each policy was evaluated three times per goal aneurysm in each of the three rearranged and three novel geometries (18 trials).} 
\label{fig:ablatedist}
\end{center}
\vspace{-1\baselineskip}
\end{figure}
\begin{table}[t]
    \caption{Success rates across each baseline on the novel geometries, including Wilson confidence intervals (CI) and pair-wise logistic regression p-values vs. Ours}
    \centering
    \begin{tabular}{c | c | c | c}
    Policy & Novel geometries & 95\% CI & \bettershortstack{Pairwise\\{}Comparison\\{}(p-value)}\\
    \bottomrule
    \toprule
    \textbf{Ours} & \textbf{83\% (15/18)} & 60-94\%  & -\\
    Diffusion~\cite{Chi2023} & 22\% (4/18) & 1-45\%  & \textless0.001\\
    MLP & 28\% (5/18) & 12-51\% & \textless0.001\\
    Centerline following & 50\% (9/18) & 29-71\% & 0.042\\
    \textbf{Clinician 1} & \textbf{83\% (15/18)} & 60-94\% & 1\\
    Clinician 2 & 78\% (14/18) & 55-91\% & 0.987\\
    \end{tabular}
    \label{tab:successbase}
\end{table}
\begin{figure}[t]
\begin{center}
\centerline{\includegraphics[width=\linewidth]{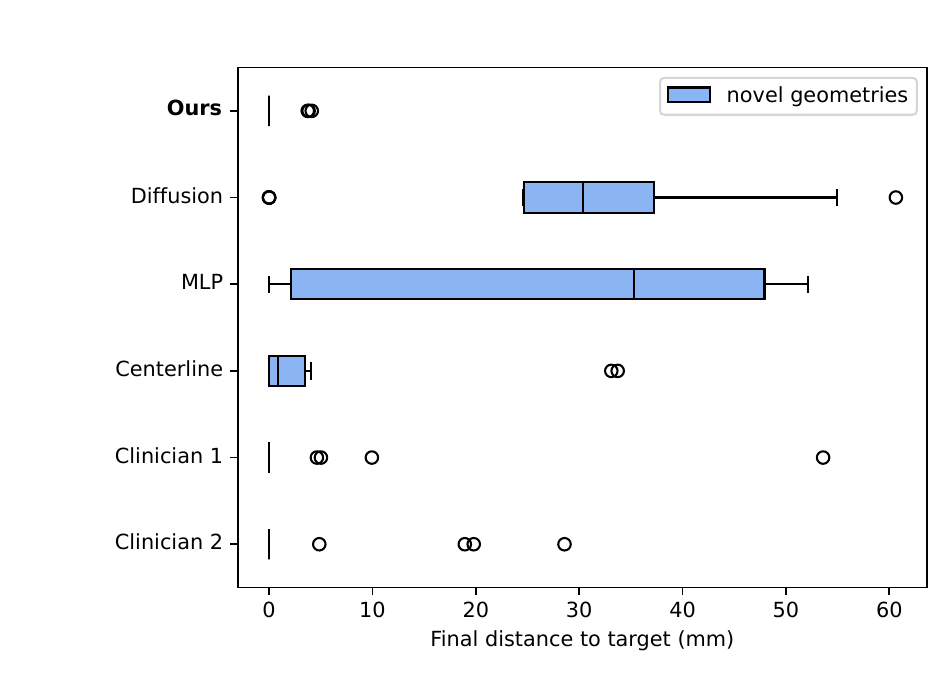}}
\caption{Distance to the goal aneurysm boundary at the end of the trial for each baseline. Each baseline was evaluated three times per goal aneurysm in each of the three novel geometries (18 trials).} 
\label{fig:basedist}
\end{center}
\vspace{-1\baselineskip}
\end{figure}
To evaluate our policy, we designed two sets of geometries that were not in the training set; (1) three rearranged geometries, which were mazes that contain blocks that appeared in the training set, but now in a unique combination, and (2) three novel geometries, which were mazes with branches and bifurcations not seen at all in the training set (see Fig.~\ref{fig:mazesetup}). On each maze, three trials per target aneurysm were attempted. The robot begins at the maze entrance, and success means that the robot's tip enters the target aneurysm within 60 seconds. The robot's tip trajectory, trial duration, and number of contrast injections were recorded. These trials were performed for ablative policies, baseline policies, and two clinicians.

We also analyzed trial-level success using binomial logistic regression, following standard methods for categorical data analysis~\cite{Agresti2007}. We performed a likelihood ratio test comparing a statistical model that used path (i.e. maze and target) and policy variant as predictors, and a model that used only path as a predictor. We found a significant overall effect of policy variant on success rate among ablations on the rearranged geometries (p\textless0.001), ablations on novel geometries (p=0.001), as well as baselines on novel geometries (p\textless0.001), indicating that policy choice explained success outcomes beyond path difficulty. Pairwise likelihood-ratio tests against the proposed policy are reported in Tables \ref{tab:successablate} and \ref{tab:successbase}, with Holm-corrected p-values. Because only six navigation paths with three trials each were evaluated per group, these analyses are interpreted as exploratory.

\textbf{Chunk size determination:} Chunk size is a key parameter to tune when applying an action-chunking imitation learning policy~\cite{Zhao2023} to a new environment. We trained a policy with chunk sizes of 1, 2, 3, and 4 sec. and evaluated them on the rearranged vascular geometries. The 2 sec. chunk size performed best (89\% success rate vs. 28\%, 61\%, and 78\% for chunk sizes of 1, 3, and 4 seconds, respectively).

\textbf{Ablations:} Next, we evaluated our key design choices: usage of recovery data, contrast injection prediction, choice of goal representation, and choice of action representation. Each ablation differed from the proposed policy only by the stated modification; success rates and final distances to the target are shown in Table~\ref{tab:successablate} and Figure~\ref{fig:ablatedist}. 

Recovery data had the clearest effect on out-of-distribution performance: removing recovery data reduced novel-geometry success from 83\% to 22\%, while using half of the recovery data reduced success to 61\%. The 50\% recovery policy performed equally well on rearranged geometries and reached a slightly lower mean final distance than the proposed policy, but this trend reversed on novel geometries, suggesting that additional recovery data is most important when local vessel shapes are unseen during training. Learned contrast injection was also important: fixed 16 sec. and 8 sec. injection intervals reduced novel-geometry success to 56\% and 50\%, respectively, despite the 8 sec. interval providing more frequent visual updates. This suggests that contrast timing encodes task context, since demonstrators often inject after mistakes or during recovery, causing fixed frequent contrast to trigger recovery-like behavior too often. Goal conditioning also affected the final interaction with the aneurysm. The binary goal policy, consisting of a binary mask with positive values inside the aneurysm circle, achieved only 61\% success but often reached close to the target (2.0$\pm$2.8 mm on novel geometries). This indicates that it lacked sufficient directional structure to reliably enter the aneurysm. Finally, absolute motor actions reduced novel-geometry success to 39\% and produced the largest rearranged-to-novel drop, consistent with the hypothesis that complex robot dynamics make absolute motor positions unreliable across trials and geometries.

\textbf{Baselines:} We evaluated several baselines as a comparison to our proposed policy: a transformer-based Diffusion policy~\cite{Chi2023}, a multi-layer perceptron (MLP), a classical centerline-following controller, and two clinicians trained in neurointerventions. The Diffusion and MLP policies use the same inputs (live and goal images passed through Resnet encoders) and outputs (action chunk and contrast prediction) as our proposed policy. The centerline-following controller used the contrast-flow BFS path and a PD controller to drive the robot tip to successive centerline points. We evaluated all these policies on the novel geometries test set. The resulting success rates are found in Table \ref{tab:successbase} and final distances to the goal in Fig. \ref{fig:basedist}, along with failure mode frequencies in Table \ref{tab:failmodestype}. 

The Diffusion policy achieved only a 22\% success rate, and often drove the robot into the maze walls. This result agrees with~\cite{Kim2024}, in which the action-chunking transformer outperforms diffusion for a series of surgical tasks.
\begin{figure}[t]
\begin{center}
\centerline{\includegraphics[width=\linewidth]{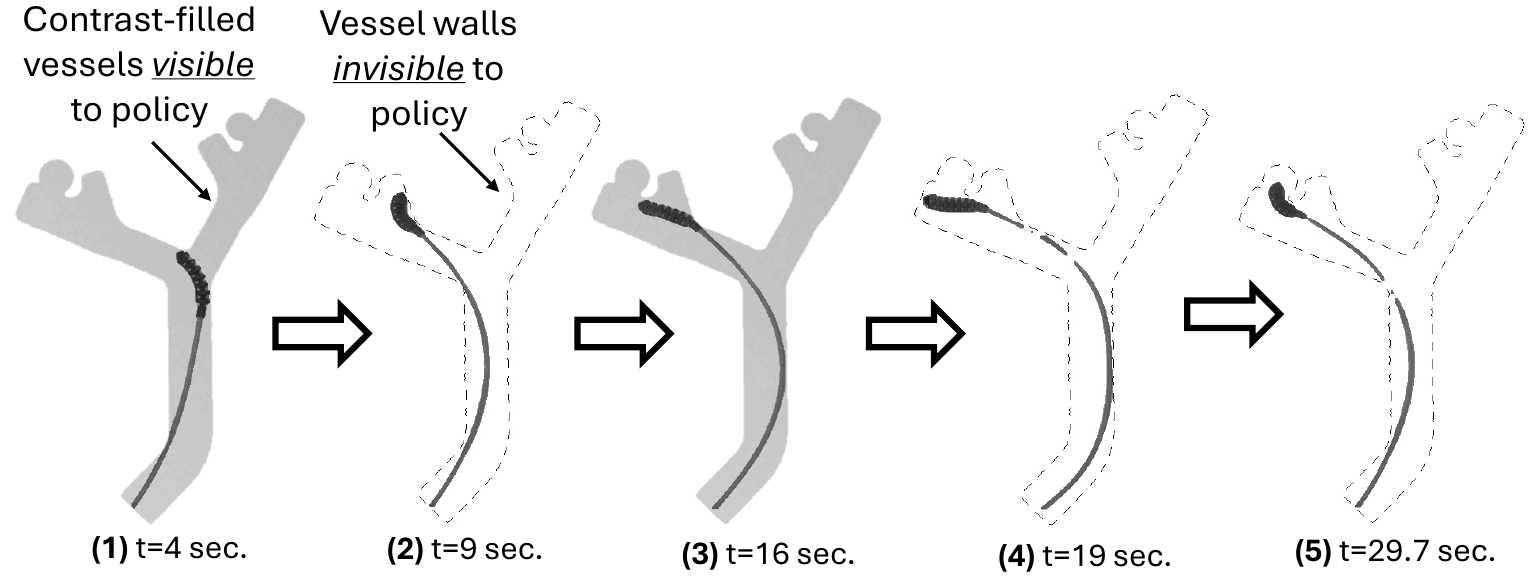}}
\caption{Representative successful rollout of our policy on a rearranged test maze. Contrast injections reveal the vessels to the policy. The policy (1) bends across the bifurcation, (2) initially enters the wrong branch, (3) recovers and advances, (4) overshoots after friction release, and (5) retracts and bends into the aneurysm.} 
\label{fig:exaamplesuccess}
\end{center}
\vspace{-1\baselineskip}
\end{figure}
The MLP achieved only a 28\% success rate and frequently became caught in an oscillatory pattern. It is postulated that the MLP's lack of temporal awareness hinders its ability to predict the multi-step maneuvers that would be required to traverse the mazes.
\begin{table}[ht]
    \caption{Failure modes of unsuccessful trials. Oscillation denotes repeated crossing of the midpoint of the final 20 sec target distance range when the range exceeded 2 mm. Stalling denotes a final 20 sec target distance range below 2 mm. Remaining failures were due to wall contact-induced buckling as confirmed by video review.}
    \centering
    \begin{tabular}{c | c | c | c}
    Policy & Oscillating & Stalling & Buckling\\
    \bottomrule
    \toprule
    Ours & 0\% (0/3) & 100\% (3/3) & 0\% (0/3)\\
    Diffusion & 29\% (4/14) & 7\% (1/14) & 64\% (9/14)\\
    MLP & 100\% (13/13) & 0\% (0/13) & 0\% (0/13)\\
    Centerline & 0\% (0/9) & 56\% (5/9) & 44\% (4/9)\\
    Clinician 1 & 100\% (3/3) & 0\% (0/3) & 0\% (0/3)\\
    Clinician 2 & 100\% (4/4) & 0\% (0/4) & 0\% (0/4)\\
    \end{tabular}
    \label{tab:failmodestype}
\end{table}
The centerline-following policy achieves a 50\% success rate despite access to perfect vessel information. The policy struggled to traverse bifurcations or recover from overshooting the target, which may require a momentary bend away from the centerline. Multi-step and trial-and-error-like movements could be explicitly encoded into the classical controller~\cite{Li2024Centerline}, but such methods have limited generalizability in comparison to learning approaches and still rely on highly accurate vessel reconstruction.

\textbf{Clinician trials:} The clinicians performed the task with the same teleoperated control handle used for demonstration collection. Clinician 1 had 25 years of experience in neurointerventions and Clinician 2 had 3 years of experience; neither had ever used the robot or mazes. Each clinician completed ten practice trials on a training maze, then performed three recorded trials per target on the novel geometries. They achieved similar success rates (Table~\ref{tab:successbase}) and final distances to the goal (Fig.~\ref{fig:basedist}) as our proposed policy. Despite similar results, the clinicians employed a more cautious strategy -- traversing the geometries slower (4.1 and 3.5 mm/sec vs. 6.3 mm/sec for ours) and using contrast less frequently (every 22.4 and 27.8 sec vs. every 7.5 sec for the proposed policy).
\begin{figure}[ht]
\begin{center}
\centerline{\includegraphics[width=\linewidth]{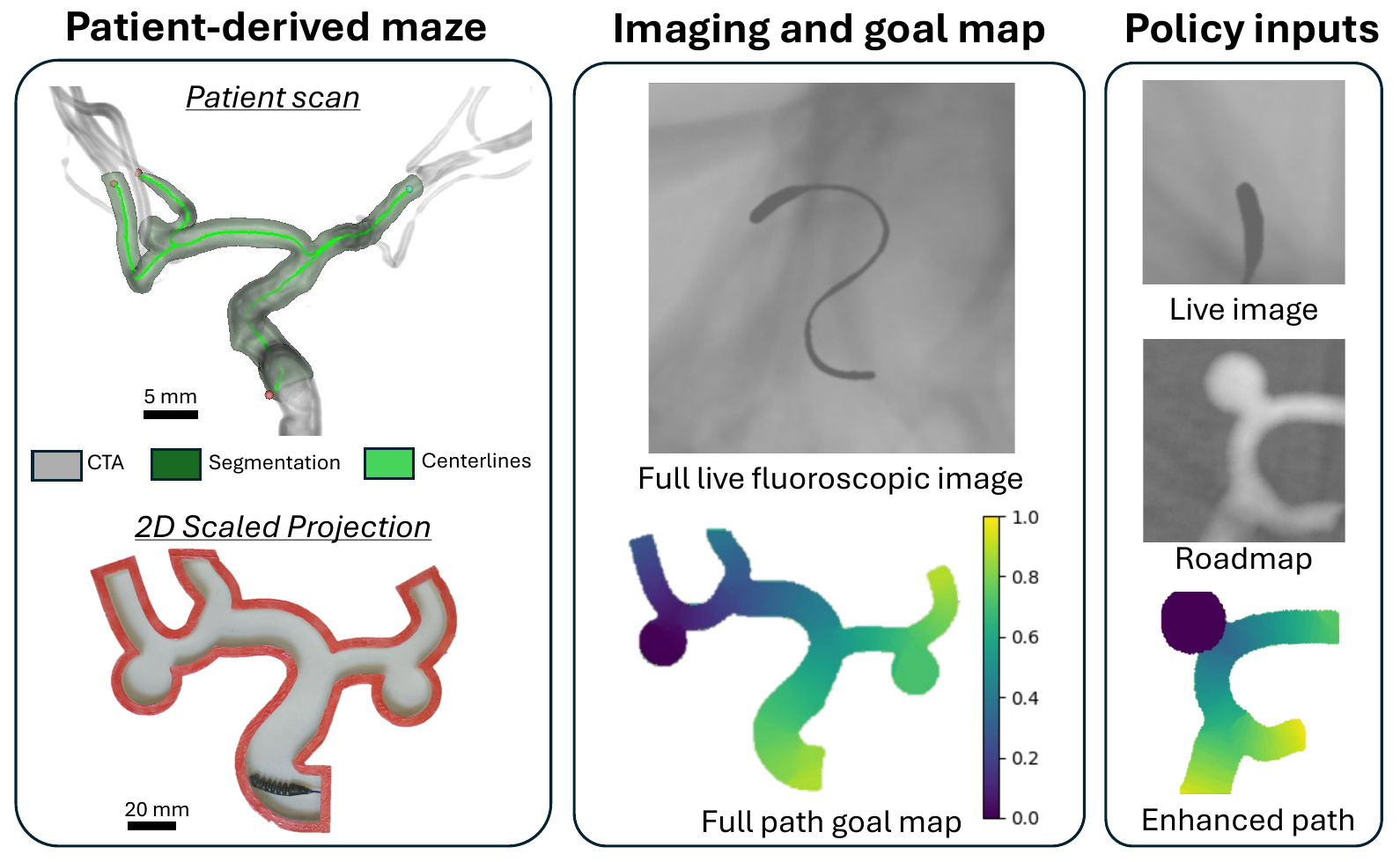}}
\caption{Adaptations to the original methods for the case of neurointervention. \textbf{(left)} A patient's CTA was transformed to a 2D maze for our robot's navigation \textbf{(middle)} Background bones, reduced image contrast, and a new goal map representation were implemented \textbf{(right)} Policy inputs were cropped around the robot's tip point to reduce the appearance gap between the original and patient-derived geometries.} 
\label{fig:neurosetup}
\end{center}
\vspace{-1\baselineskip}
\end{figure}
\section{Methods for Neurointerventional Case Study}
\label{sec:neuromethods}
\textbf{Patient-derived maze:} To evaluate whether the framework extends beyond the modular bifurcated geometries, we created a patient-derived neurovascular maze from a pre-operative computed tomography angiography (CTA). Portions of the internal carotid artery (ICA) and middle cerebral artery (MCA) were segmented, and centerlines were extracted in 3D Slicer. The 3D centerline tree was projected into a 2D plane while preserving local segment lengths and turning angles. The projected centerlines were scaled by a factor of five and shelled to create a hollow vessel structure, and two aneurysms were added. To accommodate the increased tortuosity of this maze, the robot-tube interface was mechanically reinforced, and tube stiffness was adjusted until both targets could be reached reliably under teleoperation.

\textbf{Neurovascular fluoroscopy simulation:} The fluoroscopy simulator was modified to better approximate neurointerventional imaging. From a set of ten anonymized DSA runs in the ICAs of two different patients, we obtained vessel contrast-to-noise ratios (CNRs) of 11-43, catheter CNRs of 60-75, and two-second contrast injections including a half-second rise and fall. Additionally, a section of a head x-ray was placed in the background, and the live image was set to oscillate at 70 beats per minute~\cite{ostchega2011resting} in a random direction with an amplitude of 1 mm~\cite{almudayni2023magnetic}. The resulting image appearance is shown in Fig.~\ref{fig:neurosetup}.

\textbf{Policy adaptation:} The original training demonstrations were post-processed with the neurovascular fluoroscopy parameters described above, and a new policy was trained with the following adaptations. Under the higher tortuosity of the patient-derived maze, the original ``x-y'' goal map may produce imprecise guiding vector directions. Thus, we explored centerline-based goal maps: a ``path'' goal map encoding distance-to-target along the centerline and an ``enhanced path'' goal map which adds a discrete target circle to preserve a strong local signal near the aneurysm. Since centerlines are more reliably extracted from pre-operative CTA rather than DSA, the new goal maps were shifted by a random rigid transformation with respect to the live view, with a sampling range of 10 mm and 5 degrees, to model DSA-to-CTA registration uncertainty~\cite{van2026georeg}. The policy inputs were then cropped around the tracked robot tip and rotated such that the local goal-map centerline closest to the robot tip was facing forward, resembling navigation through one vessel segment or bifurcation at a time. Figure~\ref{fig:neurosetup} depicts the updated goal maps and cropped policy inputs. Lastly, with the aim of aligning the policy's behavior better with that of the clinicians, we decreased the BCE loss weight, $\lambda_c$, from $0.5$ to $0.3$, thereby reducing contrast usage.
\section{Neurointerventional Case Study Results}
\label{sec:neuroresults}
\begin{table}[ht]
    \caption{Results for the neurointerventional case study.}
    \centering
    \begin{tabularx}{\columnwidth}{c|c|c|c|c}
    & \multicolumn{2}{c|}{Success rate} & \multicolumn{2}{c}{\bettershortstack{Median final distance\\{}[IQR] (mm)}}\\
    \midrule
    Goal map & Left & Right & Left & Right\\
    \bottomrule
    \toprule
    X-Y & 30\% (3/10) & 50\% (5/10) & \bettershortstack{38.6\\{}[5.2-52.0]} & \bettershortstack{3.9\\{}[0.1-9.1]}\\
    Path & \textbf{70\% (7/10)} & 30\% (3/10) & \bettershortstack{\textbf{0.0}\\{}[0.0-20.5]} & \bettershortstack{8.1\\{}[2.0-32.1]}\\
    Enhanced path & \textbf{70\% (7/10)} & \textbf{80\% (8/10)} & \bettershortstack{\textbf{0.0}\\{}[0.0-40.1]} & \bettershortstack{\textbf{0.4}\\{}[0.0-0.7]}
    \end{tabularx}
    \label{tab:neuroresults}
\end{table}
Three policies, one per proposed goal map (``x-y,'' ``path,'' and ``enhanced path,'') were evaluated by 10 rollouts per aneurysm target. The ``enhanced path'' policy achieved the highest overall success rate of 75\% (Table~\ref{tab:neuroresults}). The ``path'' policy achieved comparable success on the left target, but often skipped past the right target and stalled nearby. The ``x-y'' policy exhibited similar right-target failure modes and generally failed to traverse the first high-curvature bifurcation leading to the left target. The ``enhanced path'' policy averaged 14.3 sec/injection, lower than the clinicians’ 22.4 and 27.8 sec/injection, but improved over the original policy’s 7.5 sec/injection.
 \section{Discussion and Conclusion}
\label{sec:conclusion}
This work demonstrates that action-chunking imitation learning can enable autonomous soft robotic endovascular navigation. The policy reached the target aneurysms with an 83\% success rate on unseen geometries after just 7 hours of demonstration data, and the adapted policy achieved 75\% success in the more difficult neurointerventional case study. Ablations and baselines validated each design element under partial observability and complex robot dynamics. These results support the broader strategy of training primarily in bench-top synthetic vessel models before translating to real animals and humans with limited additional demonstrations.

In future work, several key improvements could be made. Currently, the proposed policy occasionally stalls near the target and uses contrast more frequently than clinicians. To address this, finer-grained ablation studies, contrast-aware objectives, and incorporating observation history~\cite{Madlekar2021} should be explored. This ``low-level'' policy should also be integrated into a hierarchical scheme for decision-making and high-level strategy~\cite{Kim2025}. Future studies should evaluate smaller, clinically relevant tools~\cite{felixsoromicrocath} in 3D vessel models. We will also explore fluoroscopic scanner positioning control, reduced reliance on preoperative information, demonstration data collected from expert neurointerventional surgeons, and testing more geometries to support stronger statistical analysis. Expert DAgger correction can remain scalable by replicating observed autonomous failure modes offline and focusing expert effort on recovery states rather than full supervised rollouts.

Altogether, these insights provide an important step toward realizing the benefits of end-to-end imitation learning for difficult endovascular procedures.
\bibliographystyle{IEEEtran}
\bibliography{IEEEabrv,refs}
\end{document}